\pdfoutput=1
\documentclass{article}
\usepackage{ijcai22}

\usepackage{times}
\usepackage{soul}
\usepackage{url}
\usepackage[hidelinks]{hyperref}
\usepackage[utf8]{inputenc}
\usepackage[small]{caption}
\usepackage{bm}
\usepackage{color}
\usepackage{graphicx}
\usepackage{amsmath}
\usepackage{amsthm}
\usepackage{booktabs}
\usepackage{algorithm}
\usepackage{algorithmic}
\usepackage{multirow}
\usepackage{subfigure}
\title{Multi-Level Attention for Unsupervised Person Re-Identification}

\author{
    Yi Zheng
    \affiliations
    China University of Mining and Technology    
    \emails
    tb19170008b2@cumt.edu.cn
}

\begin{document}

\maketitle

\begin{abstract}
    The attention mechanism is widely used in deep learning because of its excellent performance in neural networks without introducing additional information.
    However, in unsupervised person re-identification, the attention module represented by multi-headed self-attention suffers from attention spreading in the condition of non-ground truth.
    To solve this problem, we design pixel-level attention module to provide constraints for multi-headed self-attention. 
    Meanwhile, for the trait that the identification targets of person re-identification data are all pedestrians in the samples, we design domain-level attention module to provide more comprehensive \textbf{pedestrian} features.
    We combine head-level, pixel-level and domain-level attention to propose multi-level attention block and validate its performance on three 
    large person re-identification datasets (Market-1501, DukeMTMC-reID and MSMT17
    ).
\end{abstract}

\section{Introduction}

\begin{figure*}[ht]
    \centering
    \includegraphics[width=\textwidth]{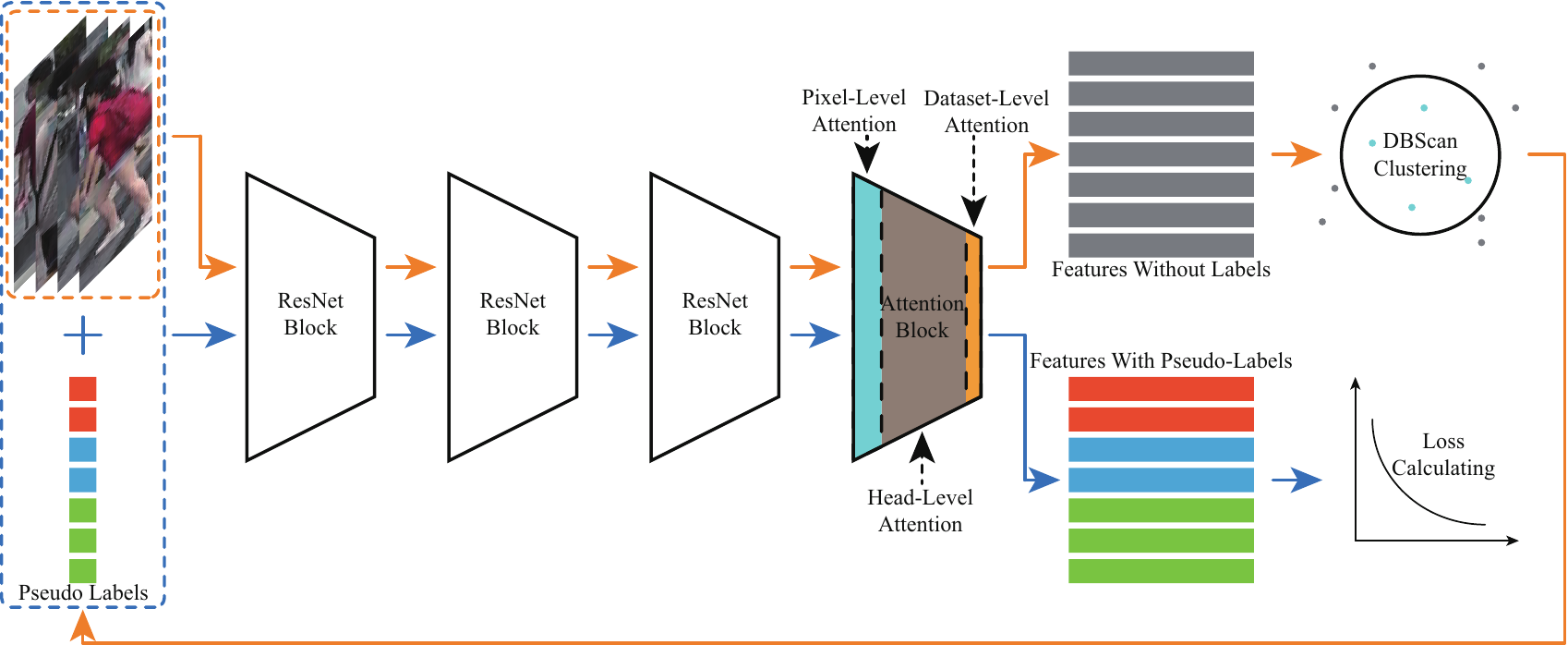}
    \caption{The training strategies of our paper. The orange arrows indicate the clustering process and the blue arrows indicate the training process of the neural network. (Please view in color.)}
    \label{work_flow}
\end{figure*}

Person re-identification faces a heavy label dependency problem like many other deep learning problems.
The high annotation cost also limits the development of existing person re-identification methods. 
Therefore, semi-supervised/unsupervised deep learning methods are attracted by more and more researchers as they do not rely too much on data annotation.
A common semi-supervised approach is to pre-train the model supervised on a labeled dataset, then enhance the generalization of the network in some ways, and transfer the network to a dataset without labels for unsupervised learning \cite{yu2019unsupervised,zhong2019invariance}. 
Usually, these approaches are also called unsupervised domain adaptation (UDA) for person re-identification.
Another fully unsupervised learning (FUL) approaches \cite{ge2020self,lin2019bottom,guo2021beyond,wang2020unsupervised} 
means that they do not use any annotated information, and the features extracted by the neural network are used to obtain pseudo-labels by clustering algorithms (such as K-means \cite{ji2020attention} and hierarchical clustering algorithm \cite{lin2019bottom}), then the pseudo-labels guide the network to mine more accurate pedestrian features.
In general, the UDA methods perform better than the FUL methods due to the introduced supervised information on the source datasets.
However, the UDA approaches are still limited by complex training processes and unignorable differences between the source and target domains, which prevent them from being a universal solution.

As a sub-problem of image retrieval, the essence of person re-identification is how accurately pedestrian features are matched. 
Therefore, how to get a neural network that can extract more accurate pedestrian features is one of the keys to deep learning-based person re-identification methods.
In supervised learning, relying on the constraint of pedestrian labels, researchers have designed methods such as part features \cite{
guo2019beyond} and attention mechanism \cite{
zhang2021rest,he2021transreid} to extract more accurate pedestrian features. 
Since the multi-headed self-attentive models represented by transformer\cite{vaswani2017attention} and VIT\cite{dosovitskiy2020image} have been proposed, the self-attentive mechanism has been accepted by more and more researchers as an effective method for extracting more discriminative features without introducing manual partition and additional annotated information.
However, due to the specificity of person re-identification data, pedestrian images captured by same camera may often contain similar backgrounds. 
Without the label constraint, it leads the unsupervised model with attention module to treat the background information as classification basis incorrectly, which affects the differentiation of pedestrian features and seriously reduces the accuracy of re-identification.

To address the above problems, we design a Multi-Level Attention (MLA) structure based on cluster contrast for unsupervised person re-identification. 
Specifically, we design an attention structure that combines three different scales: pixel-level, head-level, and domain-level. 
Head-Level Attention (HLA), which is known as Multi-Headed Self-Attention (MHSA), they are usually used to process large scale images.
And on the small-sized pedestrian images, the normal position encoding cannot well divide the image, meanwhile, because of the specificity of the pedestrian images, HLA cannot get the effective attention region.
Therefore, before the features input into HLA, we use pixel-level attention (PLA) to obtain weights of each pixel with its surrounding pixels and use them to constrain HLA to limit the attention region on the pedestrians themselves.
Considering the specificity of pedestrian data, where all targets are \textbf{ pedestrians}, we use domain-level attention (DLA) to fuse pedestrian features across identities to better distinguish between foreground (pedestrian) and background of images, which supplements image-level (both pixel-level and header-level) attention.


Overall, the MLA can optimize the extracted features and improve the clustering results; on the other hand, better clustering results can also guide the model to extract more accurate pedestrian features.
Eventually, our method achieves competitive results compared to other state-of-the-art unsupervised person re-identification models.
The following contributions are included in our paper.
\begin{itemize}
    \item We constrain the multi-head self-attention module by adding a Pixel-Level Attention (PLA) module to make the Head-Level Attention (HLA) module applicable to person re-identification data.
    \item We design and add a Domain-Level Attention (DLA) module for the whole dataset to improve the effectiveness of PLA and HLA by using the specificity of person re-identification data.
    \item With a combination of pixel-level, head-level, and domain-level attention, we successfully applied the attention module to unsupervised pedestrian re-identification task and achieved competitive results.
\end{itemize}


\section{Related Works}\label{related_works}

\subsection{Unsupervised person re-identification}

Unsupervised person re-identification based on deep learning is mainly divided into two types, unsupervised domain adaptation (UDA) methods \cite{zheng2019joint,luo2020generalizing,bai2021unsupervised} based on transfer learning  and fully unsupervised learning (FUL) methods \cite{lin2019bottom} based on clustering algorithms.
Fan et al. \cite{fan2018unsupervised} used the K-means algorithm for clustering, and they raised the confidence limit of the clustering similarity considering that the features extracted by the model were coarse at the early stage of training.
Lin et al. \cite{lin2019bottom} use hierarchical clustering, alternating the training and clustering steps, combining a fixed proportion of images into clusters at each iteration and constraining the distribution of clusters in some way.
Dai et al. \cite{dai2021cluster} also used an iterative approach, but used the DBScan algorithm for clustering and improved the update strategy of the memory dictionary during loss calculation.

\subsection{Transformer and attention mechanism}

Before the transformer structure \cite{vaswani2017attention} was proposed, many person re-identification methods consciously used attention mechanisms to obtain more discriminative features.
Zheng et al. \cite{zheng2021siamese} trained an additional affine transformation module to actively crop the pedestrian region of interest.
Zhou et al. \cite{zhou2020interpretable} proposed an interpretable attention-based part model, learning the correlation among pedestrian parts.

While ViT \cite{dosovitskiy2020image} enables transformer to handle image input, ResT \cite{zhang2021rest} frees transformer from the limitation of input image size and makes it more flexible to work with arbitrary size images.
The proposal of external attention \cite{guo2021beyond} provides the idea of using the prior knowledge of the data.
TransReID \cite{he2021transreid} applies the complete transformer structure to person re-identification networks for the first time and achieves state-of-the-art performance through the explicit position encoding.
Although the transformer and attention mechanism have achieved encouraging results in person re-identification tasks, most of the person re-identification networks using the attention module are based on supervised training data.
It means that data labels are necessary for the model training.
Currently, to the best of our knowledge, only Ji et al. \cite{ji2020attention} have designed an attention module that combines spatial attention and channel attention and achieved state-of-the-art person re-identification performance with attention mechanism under unsupervised conditions.

\section{Method}\label{method}

\begin{table}[t]
    \centering
    \begin{tabular}{|l|c|c|}
    \hline
    Datasets & Market-1501 & DukeMTMC-reID \\\hline
    without attention & 31.7\% & 23.6\% \\\hline
    with attention & 31.1\% & 22.5\%\\
    \hline
    \end{tabular}
    \caption{\label{hierarchical}Result of mAP on hierarchical clustering-based unsupervised method with and without attention module.}
\end{table}

Attention-based methods rely on the constraint of labels to guide the neural network to pay more attention to regions of the image that have distinguishing features. 
While unsupervised learning happens to lack the necessary label information, therefore to embed the self-attention module under the unsupervised learning condition requires the high performance of the baseline method itself. 
We added the attention module based on transformer structure to the hierarchical clustering-based model with low accuracy, and the results are shown in Table \ref{hierarchical}, where the performance decreases rather than increases on these datasets.


We illustrate the training strategy proposed in the paper as Figure \ref{work_flow}.
First, as the orange arrow indicates, the unlabeled images are passed through the neural network to obtain the unlabeled features. Then the pseudo-labels are obtained by clustering using the DBScan algorithm, as shown by the blue arrows, and the loss function is calculated by the constraints of the pseudo-labels to optimize the parameters of the neural network. 
In particular, we modify the last ResBlock of the ordinary ResNet by using three attention modules with different scales instead of the original convolutional layers, and their specific structures are detailed in Sections \ref{HLA}, \ref{PLA} and \ref{DLA}.

The DBScan algorithm \cite{ester1996density} requires two manually set parameters: the clustering radius $\epsilon$ and the minimum cluster size $MinPts$. In our experiments, we set $\epsilon$ to $0.4$ and $MinPts$ to $4$ to achieve the best results.
In the training phase, referring to the method proposed in the paper \cite{dai2021cluster}, the classification weight matrix is replaced by a memory dictionary, which is initialized by randomly selected sample features from each cluster and updated by the strategy of $batch\_hard$ during training.
Finally, we use the following ClusterNCE Loss \cite{dai2021cluster} as the loss function.

\begin{equation}
    L_{CNCE}=-\log \frac{\exp \left(x \cdot C_{i} / \tau\right)}{\sum_{i=0}^{K} \exp \left(x \cdot C_{i} / \tau\right)}
\end{equation}
where the feature to be classified is $x$ and the temperature parameter $\tau$ is used to control the interval between clusters. The smallest loss value is obtained if $x$ is most similar to the $i$-th cluster feature $C_i$ in the memory dictionary among the total $K$ identities.

\subsection{Head-level attention}\label{HLA}

\begin{figure}[t]
    \centering
    \includegraphics[width=0.8\linewidth]{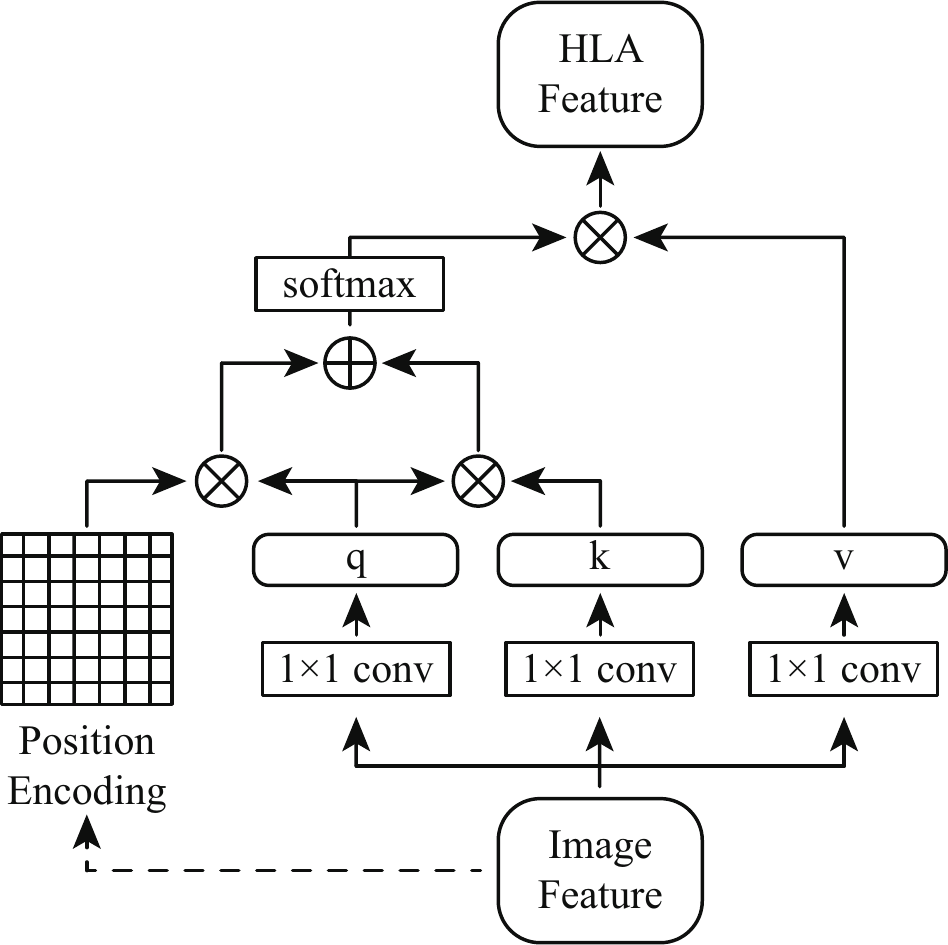}
    \caption{The Head-Level Attention (HLA) structure is the main component of the attention block. where the position encoding module is a learnable parameter and $\bigotimes$ and $\bigoplus$ denote matrix multiplication and element-wise summation, respectively. The dotted line represents the effect of the pixel-level attention module on the position encoding.}
    \label{head_attention}
\end{figure}

While the transformer module has proved its great performance on various machine learning tasks, it is also widely known for its huge memory cost, especially in computer vision. 
Generally, in the case of insufficient memory space, the batch size of training data needs to be sacrificed to accommodate large and complex models.
However, due to the storage and update requirements of the dynamic memory dictionary, the batch size in training must guarantee a certain size to select effective batch hard samples. 
BoTNet \cite{srinivas2021bottleneck} provides us with an effective solution.
Srinivas et al. argue that replacing the convolutional layer with a transformer structure on only the last residual block close to the semantic features can improve the performance of the whole model without significantly increasing the network parameters.

As shown in Figure \ref{head_attention}, we use three $1 \times 1$ convolutional layers as $W_q$, $W_k$ and $W_v$ to obtain $q$ (query), $k$ (key) and $v$ (value) respectively.
In which, $q$ is matrix multiplied with $k$ to obtain the weights corresponding to $v$.
Meanwhile, in VIT \cite{dosovitskiy2020image}, position embedding is manually assigned to each patch, similarly, we multiply position encoding with $q$ to obtain this similar correspondence.
To complete the embedding of the position encoding, we sum this correspondence with the weight information ($q \bigotimes k$) element-wisely. 
After softmax, the weight information containing the position encoding is matrix multiplied with $v$ to output the final head-level attention features.

\subsection{Pixel-level attention}\label{PLA}

\begin{figure}[!t]
    \centering
    \includegraphics[width=0.5\linewidth]{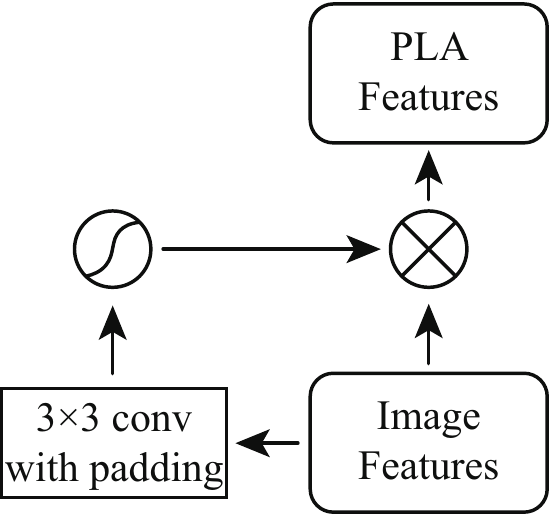}
    \caption{The structure of Pixel-Level Attention (PLA).}
    \label{pix_attention}
\end{figure}

In BoTNet \cite{srinivas2021bottleneck}, the position encoding $pos$ is obtained by summing two matrices obeying standard normal distribution element-wisely.
This position encoding achieves global ($all2all$) self-attention on the 2-D feature map, but it is not suitable for the person re-identification. 
Compared with multi-target classification, pedestrian images usually contain only one target to be classified and the size of the image is small, and all the backgrounds belong to distracting information.
Meanwhile, pedestrian images captured by the same camera contain similar backgrounds, so under unsupervised conditions, $all2all$ attention will make the network pay more attention to image regions unrelated to pedestrians and use this distracting information as the clustering basis, leading to worse clustering results.
Therefore, we believe that there must be an initial weight assignment between the foreground and background of the pedestrian image in the features before the position encoding information is embedded into the features.
Considering the baseline model already has a good performance, we consider that the original features already contain the initial distinction between foreground and background.
Therefore, by assigning weights pixel-by-pixel to the image features before inputting them into the head-level attention module, we generate a new feature map that stores attention information with pixel-level precision, which we call pixel-level attention (PLA).

First, let us review the ViT \cite{dosovitskiy2020image} model, where the input image is divided into $N$ tokens, and a variable is added to each token to encode the position. 
When the input token is $x$ and the position is encoded as $E_{pos}$, the input with position encoding $\hat{x}$ can be expressed as:

\begin{equation}
    \hat{x}=x+E_{pos}
\end{equation}
where $x \in N$ and $E_{pos} \in N$. However, these position labels are built on the division of the whole input image, and the size and number of divisions are fixed.
For pedestrian images, this division obviously has great limitations.
Therefore, we divide the image by pixels, each token $x$ represents one feature pixel and replace the position encoding with weight, then the above equation can also be seen as pixel-wise attention which is encoded by weights.
In general, we want to get the weight of a pixel on a 2D feature map, but at the same time we do not want to get the global weights, but only the weight of a small area around this pixel.
Combining the above requirements, it is obvious that convolution is a very useful and effective method. 
Specifically, as shown in Figure \ref{pix_attention}, for a pixel $x$ on the 2-D feature map, we can obtain the pixel-wise weights by using a $3\times3$ convolution layer with zero-padding and use sigmoid ($\sigma(\cdot)$) for scaling the weights. Then the weights are no longer added to the pixel $x$ but multiplied with it. 
Therefore we can represent the pixel attention PLA as follows:

\begin{equation}
    \operatorname{PLA}(x)=x \cdot \sigma\left(\operatorname{Conv}_{\text {padding }}^{3 \times 3}(x)\right)
\end{equation}

\begin{figure}[!t]
    \centering
    \includegraphics[width=0.23\linewidth]{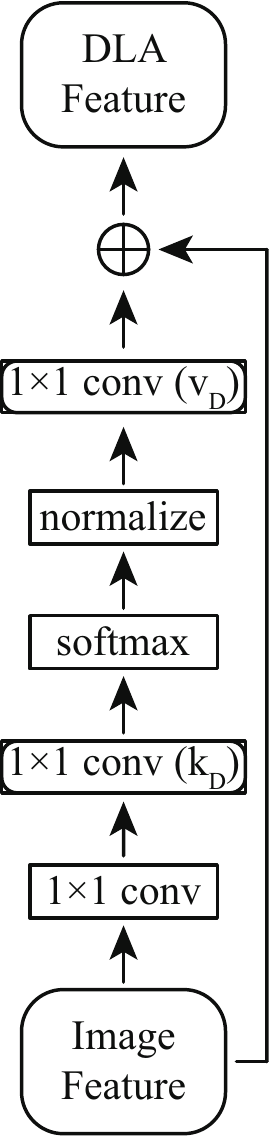}
    \caption{The structure of Domain-Level Attention (DLA).}
    \label{dataset_attention}
\end{figure}

\subsection{Domain-level attention}\label{DLA}

The person re-identification task differs from the conventional classification task in that even pedestrians with different identities share the general features of the target \textbf{pedestrian} over the whole dataset.
Therefore, we believe that training an attention module over the entire dataset can combine the \textbf{pedestrian information} in all images.
On the one hand, the learned attention enables the network to better distinguish between the foreground and background of pedestrian images and to be more robust to the complex shift of angles in pedestrian data; on the other hand, it enables the network to focus on more complete pedestrian information that is generalized across the entire dataset, rather than just a particular part of the pedestrian.

As shown in Figure \ref{dataset_attention}, for the input feature $F^{h\times w\times c}$, where $h$ and $w$ are the length and width of the feature, respectively, and the number of channels is $c$.
Similar to the way in HLA, we obtain $q$ through a convolutional layer with the kernel of $1\times1$.
After that, we design two $1\times1$ convolution layers without bias as storage units, representing the data set level $k_D^{1\times1}$ and $v_D^{1\times1}$, for replacing $k^T$ and $v$ in self-attention. 
where the number of (input channel, output channel) of $k_D^{1\times1}$ and $v_D^{1\times1}$ are $\left(c,c_k\right)$ and $\left(c_k,c\right)$, respectively, and their initial weight parameters are transposed to each other. 
In addition, we add an extra step of normalization after softmax, which can reduce the effect of outlier features (such as features of excessively obscured pedestrian samples) and prevent attention from failing.
Finally, the whole process of domain-level attention can be described as:

\begin{equation}
    q^{h \times w \times c}=\operatorname{conv}^{1 \times 1}\left(F^{h \times w \times c}\right)
\end{equation}

\begin{equation}
    q k^{T}=\operatorname{softmax}\left(k_{D}^{1 \times 1 \times c_{k}}\left(q^{h \times w \times c}\right)\right)
\end{equation}

\begin{equation}
    \begin{split}
        \operatorname{DLA}\left(F^{h \times w \times c}\right) &= q k^{T} v\\
        &= F^{h \times w \times c}+v_{D}^{1 \times 1 \times c}\left(\operatorname{Norm}\left(q k^{T}\right)\right)
    \end{split}
\end{equation}
where $Norm$ represents the normalization operation.

\section{Experimental Results}\label{experiment}

\subsection{Datasets and implementation details}

We validate our proposed method on three well-known large-scale real person re-identification datasets (Market-1501 \cite{zheng2015scalable}, DukeMTMC-reID \cite{ristani2016performance} and MSMT17 \cite{wei2018person}).

We use ResNet-50 \cite{he2016deep} which is pre-trained on ImageNet \cite{deng2009imagenet} as the backbone model and use the DBScan algorithm \cite{ester1996density} as the clustering method as mentioned before.

Specifically, we tested our approach in an experimental environment with two 1080ti GPUs and 32G RAM. The size of the input images is set to $256 \times 128$, and each batch consists of 64 images, including 8 pseudo-identities each of which contains 8 images.
As mentioned before, the two parameters $\epsilon$ and $MinPts$ required by DBScan were set to $0.4$ and $4$, respectively, and the rest of the settings were the same as in \cite{ge2020self}. 
The number of clustering iterations is 50, the learning rate is initialized with 1.6e-4, and decreases by 90\% for every 20 epochs.

\subsection{Comparison with state-of-the-art methods}

\begin{table}[!t]
    \centering
    \begin{tabular}{l|cccc}
    \hline
    \multirow{2}{*}{Methods} & \multicolumn{4}{c}{Market-1501}\\
    \cline{2-5}
    ~ & mAP  & top-1 & top-5 & top-10  \\ 
    \hline
    BUC \shortcite{lin2019bottom} &  38.3 & 66.2  & 79.6  & 84.5    \\ 
    SSL \shortcite{lin2020unsupervised} &  37.8 & 71.7  & 83.8  & 87.4 \\
    MMCL \shortcite{wang2020unsupervised} &  45.5 & 80.3  & 89.4  & 92.3    \\
    HCT \shortcite{zeng2020hierarchical} &  56.4 & 80.0  & 91.6  & 95.2    \\
    *MMCL \shortcite{wang2020unsupervised} &  60.4 &84.4 &92.8 &95.0 \\
    CycAs \shortcite{wang2020cycas} &64.8 &84.8 &- &- \\
    UGA \shortcite{wu2019unsupervised} &70.3 &87.2 &- &- \\
    SPCL \shortcite{ge2020self}  &  73.1 & 88.1  & 95.1  & 97.0    \\
    *MMT \shortcite{ge2020mutual} & 75.6 &89.3 &95.8 &97.5 \\
    *SPCL \shortcite{ge2020self}  &  77.5 &89.7 &96.1 &97.6    \\
    CCR \shortcite{dai2021cluster} &  80.6 & 91.5  & 96.8  & 97.9    \\
    \textbf{Ours}          & \textbf{83.1} & \textbf{92.8}  & \textbf{97.1}  & \textbf{98.0}   \\
    \hline
    \end{tabular}
    \caption{Experiments on Market-1501 dataset. * means UDA methods}
    \label{market-1501}
\end{table}

\begin{table}[!t]
    \centering
    \begin{tabular}{l|cccc}
        \hline
        \multirow{2}{*}{Methods} & \multicolumn{4}{c}{DukeMTMC-reID}\\
        \cline{2-5}
        ~ & mAP  & top-1 & top-5 & top-10  \\ 
        \hline
        BUC \shortcite{lin2019bottom} &  27.5 & 47.4  &  62.6 & 68.4  \\ 
        SSL \shortcite{lin2020unsupervised} &  28.6 &52.5& 63.5& 68.9 \\
        HCT \shortcite{zeng2020hierarchical} &  50.7 &69.6 &83.4 &87.4 \\
        *MMCL \shortcite{wang2020unsupervised} &  51.4 &72.4 &82.9 &85.0 \\
        UGA \shortcite{wu2019unsupervised} &53.3 &75.0 &- &- \\
        CycAs \shortcite{wang2020cycas} &60.1 &77.9 &- &- \\
        *MMT \shortcite{ge2020mutual} & 65.1 &78.9 &88.8 &92.5\\
        SPCL \shortcite{ge2020self}  &  65.3 &81.2 &90.3 &92.2    \\
        *SPCL \shortcite{ge2020self}  & 68.8 &82.9 &90.1 &92.5 \\
        CCR \shortcite{dai2021cluster} &  69.5 &83.3 &90.4 &92.8 \\
        \textbf{Ours}  & \textbf{71.8} & \textbf{84.3}  & \textbf{91.5}  & \textbf{93.2}   \\
        \hline
    \end{tabular}
    \caption{Experiments on DukeMTMC-reID dataset. * means UDA methods}
    \label{duke}
\end{table}

\begin{table}[!ht]
    \centering
    \begin{tabular}{l|cccc}
        \hline
        \multirow{2}{*}{Methods} & \multicolumn{4}{c}{MSMT17}\\
        \cline{2-5}
        ~ & mAP  & top-1 & top-5 & top-10  \\ 
        \hline
        *ECN \shortcite{zhong2019invariance}& 10.2 &30.2 &41.5 &46.8 \\
        MMCL \shortcite{wang2020unsupervised} &  11.2 &35.4 &44.8 &49.8 \\
        TAUDL \shortcite{li2018unsupervised}& 12.5 &28.4 &- &-\\
        SPCL \shortcite{ge2020self}  &  19.1 &42.3 &55.6 &61.2    \\
        UGA \shortcite{wu2019unsupervised} &21.7 &49.5&- &-\\
        *MMT \shortcite{ge2020mutual} &24.0 &50.1 &63.5 &69.3 \\
        CycAs \shortcite{wang2020cycas} &26.7 &50.1 &- &- \\
        *SPCL \shortcite{ge2020self}  &26.8 &53.7 &65.0 &69.8 \\
        CCR \shortcite{dai2021cluster} &  31.4 & 61.2 & 72.3 & 76.4 \\
        \textbf{Ours} & \textbf{35.6} & \textbf{63.8} & \textbf{75.3}  & \textbf{79.5}\\
        \hline
    \end{tabular}
    \caption{Experiments on MSMT17 dataset. * means UDA methods}
    \label{msmt17}
\end{table}

\begin{table}[!t]
    \centering
    \begin{tabular}{l|cccc}
        \hline
        \multirow{2}{*}{Methods} & \multicolumn{4}{c}{PersonX}\\
        \cline{2-5}
        ~ & mAP  & top-1 & top-5 & top-10  \\ 
        \hline
        *MMT \shortcite{ge2020mutual} &78.9 &90.6 &96.8 &98.2 \\
        SPCL \shortcite{ge2020self}  & 78.5 &91.1 &97.8 &99.0    \\
        *SPCL \shortcite{ge2020self}  &72.3 &88.1 &96.6 &98.3 \\
        CCR \shortcite{dai2021cluster} & 84.8 &94.5 &98.4 &99.2 \\
        \textbf{Ours}          & \textbf{86.5} & \textbf{94.8}  & \textbf{98.6}  & \textbf{99.5}   \\
        \hline
        \end{tabular}
    \caption{Experiments on PersonX dataset. * means UDA methods}
    \label{personx}
\end{table}


In Table \ref{market-1501}, \ref{duke} and \ref{msmt17}, we compare the results of our method with other FUL and UDA methods. On three datasets, our method achieved the state-of-the-art results. With the same experimental equipment, besides top-n accuracy, which is higher than CCR \cite{dai2021cluster} on all datasets, our method achieves at least \textbf{2.3\%} improvement in mAP over CCR especially on the more complex MSMT17 dataset, which reaches \textbf{4.2\%}.

\subsection{Ablation studies }

\begin{table*}[!t]
    \centering
    \begin{tabular}{l|cccc|cccc}
    \hline
         Datasets & \multicolumn{4}{c|}{Market-1501} &  \multicolumn{4}{c}{DukeMTMC-reID} \\\hline
         Methods & mAP & Rank-1 & Rank-5 & Rank-10 & mAP & Rank-1 & Rank-5 & Rank-10 \\ \hline
         baseline & 80.6  & 91.5  & 96.8  & 97.9  & 69.5  & 83.3  & 90.4  & 92.8  \\ 
         HLA & 0.9  & 2.2  & 5.1  & 6.9  & 20.0  & 34.3  & 40.8  & 43.2  \\ 
         PLA & 82.4 & 92.3 & 96.6 & 97.9 & 69.8  & 83.1  & 90.4  & 93.0  \\ 
         PLA+HLA & 79.3  & 90.6  & 96.2  & 97.4  & 69.1  & 82.9  & 90.5  & 92.1  \\ 
         DLA & 82.4  & 92.3  & 96.7  & 97.7  & 70.4  & 83.4  & 92.0  & 93.4  \\ 
         ALL & 83.1  & 92.8  & 97.1  & 98.0  & 71.8  & 84.3  & 91.5  & 93.2  \\ 
        \hline
    \end{tabular}
    \caption{Ablation experiments on Market-1501 and DukeMTMC-reID datasets between the proposed modules.}
    \label{Ab1}
\end{table*}

\begin{table*}[!t]
    \centering
    \begin{tabular}{l|cccc|cccc}
    \hline
         Datasets & \multicolumn{4}{c|}{MSMT17} &  \multicolumn{4}{c}{PersonX} \\ \hline
         Methods & mAP & Rank-1 & Rank-5 & Rank-10 & mAP & Rank-1 & Rank-5 & Rank-10 \\ \hline
         baseline & 31.4  & 61.2  & 72.3  & 76.4  & 84.8  & 94.5  & 98.4  & 99.2   \\ 
         HLA & 27.7  & 55.8  & 67.2  & 71.9  & 81.6  & 92.2  & 98.0  & 99.1   \\ 
         PLA & 32.0  & 61.6  & 72.8  & 77.2  & 86.2  & 94.8  & 98.6  & 99.4   \\ 
         PLA+HLA & 30.3  & 59.1  & 70.8  & 75.1  & 86.4  & 94.8  & 98.6  & 99.   \\ 
         DLA & 30.6  & 58.9  & 70.3  & 75.0  & 86.4  & 94.8  & 98.7  & 99.4   \\ 
         ALL & 35.6  & 63.8  & 75.3  & 79.5  & 86.5  & 94.8  & 98.6  & 99.5   \\ 
        \hline
    \end{tabular}
    \caption{Ablation experiments on MSMT17 and PersonX datasets between the proposed modules.}
    \label{Ab2}
\end{table*}


We validate the impact between our proposed three levels of attention modules on the Market-1501 and DukeMTMC-reID datasets, and use Grad-CAM \cite{selvaraju2017grad} to visualize the heat maps of the features obtained from different modules, with the original samples taken from the Market-1501 dataset.

\begin{figure*}[!t]
	\centering
	\subfigure[origin]{
		\includegraphics[]{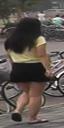}
		\label{vis_origin}
	}
   \subfigure[baseline]{
		\includegraphics[]{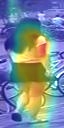}
		\label{vis_base}
	}
	\subfigure[HLA]{
		\includegraphics[]{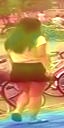}
		\label{vis_HLA}
	}
	\subfigure[PLA]{
		\includegraphics[]{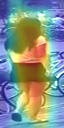}
		\label{vis_PLA}
    }
	\subfigure[PLA+HLA]{
		\includegraphics[]{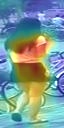}
		\label{vis_PLA_HLA}
	}
	\subfigure[DLA]{
		\includegraphics[]{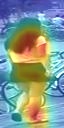}
     	\label{vis_DLA}
	}
	\subfigure[ALL]{
		\includegraphics[]{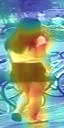}
		\label{vis_ALL}
	}
	\caption{Heat map visualization results for features of different modules.}
	\label{vis}
\end{figure*}

As shown in Table \ref{Ab1}, HLA heavily reduces the discriminability of pedestrian features in unsupervised learning without constraints of identity labels. From Figure \ref{vis_HLA}, it can be seen that, as we analyzed before, HLA cannot correctly distinguish the foreground and background of the image, except for some parts of the ground.


Comparing Figure \ref{vis_base} and Figure \ref{vis_PLA}, besides expanding the range of attention slightly, PLA improves the attention to the body parts of the pedestrians. After combining PLA and HLA, the attention in Figure \ref{vis_PLA_HLA} is more fitted to the pedestrians compared to Figure \ref{vis_HLA} and \ref{vis_PLA}.

The results in Table \ref{Ab1} show that not only the whole MLA structure can obtain better performance than the baseline method, but also adding DLA individually can significantly improve the model performance.
Meanwhile, it is clear from Figure \ref{vis_DLA} that the DLA pays attention to the pedestrian's feet in addition to the pedestrian's body. And unlike Figures \ref{vis_base}, \ref{vis_PLA} and \ref{vis_PLA_HLA}, the small portion of the background above the image is excluded. Eventually, the pedestrian description in MLA features is more comprehensive and also better distinguishes the background and foreground.


\section{Conclusion}\label{conclusion}

In this paper, we design a multi-scale attention module containing pixel-level, head-level, and domain-level attention.
Constraining multi-headed attention by inter-pixel relationships and foreground features of the dataset without introducing label information and achieving state-of-the-art performance.
In the next work, we hope to further optimize the structure of multi-level attention by extending it to the overall neural network rather than limiting it to the last residual block, referring to the pyramidal structure for a more comprehensive pedestrian feature description.









\bibliographystyle{named}
\bibliography{ijcai22}

\end{document}